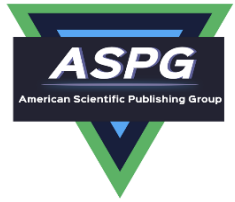

# UNCA: A Neutrosophic-Based Framework for Robust Clustering and Enhanced Data Interpretation

**D. Dhinakaran[1,*], S. Edwin Raja[1], S. Gopalakrishnan[2], D. Selvaraj[3], S. D. Lalitha[4]**

[1]Department of Computer Science and Engineering, Vel Tech Rangarajan Dr. Sagunthala R&D Institute of Science and Technology, Chennai, India

[2]Department of Computer Science & Engineering (Data Science), Madanapalle Institute of Technology & Science, Andhra Pradesh, India

[3]Department of Electronics and Communication Engineering, Panimalar Engineering College, Chennai, India

[4]Department of Computer Science and Engineering, R.M.K. Engineering College, Chennai, India

Emails: dhinaads@gmail.com; edwinrajas@gmail.com; gopalakrishnans@mits.ac.in; mails2selvaraj@yahoo.com; sdl.cse@rmkec.ac.in

**Abstract**

Accurately representing the complex linkages and inherent uncertainties included in huge datasets is still a major difficulty in the field of data clustering. We address these issues with our proposed Unified Neutrosophic Clustering Algorithm (UNCA), which combines a multifaceted strategy with Neutrosophic logic to improve clustering performance. UNCA starts with a full-fledged similarity examination via a λ-cutting matrix that filters meaningful relationships between each two points of data. Then, we initialize centroids for Neutrosophic K-Means clustering, where the membership values are based on their degrees of truth, indeterminacy and falsity. The algorithm then integrates with a dynamic network visualization and MST (Minimum Spanning Tree) so that a visual interpretation of the relationships between the clusters can be clearly represented. UNCA employs Single-Valued Neutrosophic Sets (SVNSs) to refine cluster assignments, and after fuzzifying similarity measures, guarantees a precise clustering result. The final step involves solidifying the clustering results through defuzzification methods, offering definitive cluster assignments. According to the performance evaluation results, UNCA outperforms conventional approaches in several metrics: it achieved a Silhouette Score of 0.89 on the Iris Dataset, a Davies-Bouldin Index of 0.59 on the Wine Dataset, an Adjusted Rand Index (ARI) of 0.76 on the Digits Dataset, and a Normalized Mutual Information (NMI) of 0.80 on the Customer Segmentation Dataset. These results demonstrate how UNCA enhances interpretability and resilience in addition to improving clustering accuracy when contrasted with Fuzzy C-Means (FCM), Neutrosophic C-Means (NCM), as well as Kernel Neutrosophic C-Means (KNCM). This makes UNCA a useful tool for complex data processing tasks.

## 1.　Introduction

Data clustering, a key component of data analysis is common in many areas including pattern detection, artificial intelligence, and machine learning [1]. This involves partitioning a dataset into non-overlapping clusters such that data points in the same cluster are very similar to each other while data points in other clusters are highly dissimilar [2]. An important task, which simplifies both classification and anomaly detection, as well as summarization, is finding trends and patterns found within the data. However, traditional clustering methods often meet some challenges to high-dimensional, complex data such as noise, indistinctness and the need for exact distance calculations and they are important nevertheless [3-5]. However, the long-time two-stage processing where some clustering algorithms suffer from the structure less of high dimensionality and the scale inhabited data, insufficient information generate out new advanced clustering algorithms against such as high dimensionalities problem and unclear dimensionalities criterions limited by insufficient information in them, inherent uncertainties stimulus;

Prompting clustering algorithms development in geared to ever more heaped high dimensionality dataset in new structure. Although having utility in some contexts, conventional clustering techniques (e.g., K-Means, hierarchical clustering and DBSCAN) often face these challenges [6]. K-Means, for instance, needs the number of clusters to be specified in advance and assumes the form of clusters are spherical, which might not correspond to the nature of the data. Hierarchical based approaches, while still flexible, can be computationally expensive and fail for big data. DBSCAN pays attention to density-based clustering, and can cope with noise and changing cluster density, but is sensitive to adjustments of parameters.

To overcome these limitations, complex clustering mechanism have been explored for that purpose. Partial membership is a concept presented by fuzzy clustering methods (e.g., Fuzzy C-Means (FCM)) that allows data points to belong to multiple groups with varying certainty [7]. Although doing so is useful to overcome some of the limitations of crisp clustering, it may still not be able to manage the vagueness and ambiguity inherent in real-world data. Neutrosophic logic has gained some popularity recently that provided a new perspective on clustering uncertainty handling. Neutrophilic logic provides three different membership functions over truth, indeterminacy, and falsity that fuzzy logic just covers one [8-10]. A new approach that provides a richer, more comprehensive representation of relationships among data, allowing for better consideration of uncertainty and partial truth. Neutrosophic clustering techniques exploit this enhanced logic in order to achieve higher clustering effectiveness in unclear and uncertain situations. Despite advancements in the state of the art in clustering, significant challenges persist. Traditional clustering approaches often fail to accurately reflect the data because of the inherent ambiguity present in this type of clustering, or because the cluster memberships of each data point are ambiguous therefore varying between clusters. In addition, the built-in relationships of clusters may not be clearly revealed by traditional approaches (i.e. dendrograms), making it complex to understand and visualize the results of clustering. It requires a robust clustering framework, which can handle uncertainty, assign accurate clusters, and is interpretable on complex datasets. We focus on developing a new clustering method that will be able to overcome the issues of the currently deployed clustering methods by fusing the Neutrosophic logic with the state-of-the-art clustering methods. Specifically, we aim to:

**Enhance Clustering Accuracy:** Our aim is to enhance the precision of cluster allocations when faced with ambiguity and uncertainty by implementing Neutrosophic reasoning. Creating a technique that can deal with partial membership and various levels of truth, indeterminacy, as well as falsehood is necessary to do this.

**Improve Interpretability:** We aim to provide clear insights into the relationships between clusters by using dynamic network visualization and Minimum Spanning Tree (MST) techniques. This will help in understanding the structure of the clustering results and the connections between different clusters.

**Refine Cluster Assignments:** Our objective is to improve cluster assignments by using sophisticated methods like Single-Valued Neutrosophic Sets (SVNSs) and comprehensive similarity measurements. By doing this, clusters will reflect the information's fundamental distribution more faithfully.

**Establish a Sturdy Clustering Framework:** Our goal is to provide a clustering system that can handle complicated and high-dimensional datasets with ease, while remaining resilient to noise and fluctuating cluster densities.

To fulfil these goals, in this work we propose the Unified Neutrosophic Clustering Algorithm (UNCA), a unified methodology integrating various aggregation techniques into an overall model. UNCA combines the Neutrosophic logic with the dynamic network visualization and constructs the Minimum Spanning Tree (MST) to provide improved clustering. The first step is to construct a λ-cutting matrix by setting a confidence level. Thus, this matrix can be used to facilitate the filtering of the meaningful similarities between the data points by thresholding the similarity matrix. Choosing a proper confidence level λ, the similarity relationships are further filtered to retain only the relevant relations among proteins, setting the foundation for successful clustering. After that, we perform initialization of Neutrosophic K-Means clustering algorithm by specifying the number of clusters and initializing centroids. Methods to initialize the centroids include random selection, K-Means++, heuristic approach. Neutrosophic membership values, as built on the components of Neutrosophic logic ('truth', 'indeterminacy', and 'falsity'), enable such a more sophisticated characterization of data point memberships. In this stage, Neutrosophic logic is applied to calculate membership values and centroids are updated on this basis. Without going into the detailed convergence procedure that looks at the degrees of truth, indeterminacy, and falsity to iterate toward convergence in membership vs centroids, the approach allows for iteratively refining cluster assignments.

UNCA combines dynamic network visualization with the construction of a Minimum Spanning Tree (MST) to increase interpretability. This dynamic network visualizes clusters as nodes and similarities between clusters as edges between the nodes, forming a diagram of the relationship between clusters. Constructing an MST links together clusters of nodes with the lowest aggregate similarity weight, revealing the highest impact links while also giving a view of the broader structure of the clustering. Single-Valued Neutrosophic Sets (SVNSs) are then applied to re-define similarity measures to further refine clustering results. A similarity matrix with the same scale is constructed and combined with the λ-cutting matrix to compute the new cluster assignment. This adjustment guarantees that the ultimate clustering corresponds to the true data distribution and properly addresses uncertainty. The last step is to assign data points to clusters, which have the highest possible value of the truth, which leads to fuzzy methods for determining cluster assignments through defuzzification approaches that converts the fuzzy membership values into a crisp cluster assignment. Max membership methods, weighted average and centre of gravity methods are employed to provide non-fuzzy and intuitive clustering assignments.

## 2. Related Work

Data clustering is an important step in data analysis, and given imprecise/unreliable data, this should figure even more prominently in our analyses. These types of complexity often lead to challenges for conventional clustering methods, thus needing the design of advanced methods in order to improve accuracy and stability. This paper provides a fresh perspective on the challenges in the current state of clustering approaches in the domain of Neutrosophic Clustering approaches and their contributions respectively. Acurio et al. [11] presented Neutrosophic C-Means (NCM), a new approach to resolve uncertain data through integration of fuzzy C-means and neutrosophic sets. Incorporating the concept of indeterminacy enables NCM to be an effective clustering method in ambiguous data contexts like heart disease prediction. While innovative, NCM has limitations on both scalability and interpretability of cluster boundaries. Kandasamy et al. [12] defined Double-Valued Neutrosophic Set (DVNS) and improved the methods by separating the indeterminacy into two types of indeterminate values, which one moves towards (i) truth membership and (ii) false membership. This change is intended to make the clustering results, more sensitive and precise. This approach, however, is quite complex, and it takes careful parameter tuning to achieve good clustering results.

Yuvaraj et al. [13] introduced RNCM-FSSA (RNCM with FSSA) by combining Robust Neutrosophic C-Means Clustering algorithm with Fish School Search Algorithm. They addresses the problem of task allocation to appropriate resources by clustering based on resource needs and resource supply. Although RNCM-FSSA address the indeterminacy in clustering process well, but it is a computationally expensive process, and needs some effective optimization to avoid premature convergence. Qiu et al. [14] focused on Interval-valued data that was followed by Interval Neutrosophic C-Means (INCM). INCM enhances the clustering accuracy via a new objective function and updating cluster sub prototypes iteratively. Nevertheless, it is limited to its adaptability for datasets with different characteristic and complexity. Ahmed et al. [15] proposed a framework OCE-NGC based on genetic algorithm that works towards clustering optimization of launching speculated clustering optimization estimation for the Neutrosophic Graph Cut Algorithm, with its application on skin lesion image segmentation. Though this gives the best segmentation as output, it completely depends on the best-optimized values of initial cluster centroids, which may not always give the best segmentation in diverse spaces. Hoang et al. [16] presented an approach of fuzzy clustering where the clusters were extracted through applying Neutrosophic Association Matrix. This approach proceeds by fuzzifying data into neutrosophic sets followed by lambda cutting for structure cluster definition. Though clever, this method is non-trivial, and presents serious challenges for large datasets in terms of computational expense.

Abdelhafeez et al. [17] focused on Neutrosophic C-Means Clustering (NCMC) for skin cancer detection. They have been introduced a new method which is a fusion of fuzzy C means and neutron sets, which can deal with such ambiguity of data. Despite its promise, there are limitations of NCMC, such as the difficulty of properly defining objective functions and the inability to work with different data types. Yang et al. [18] presented the Gaussian-Kernel NCM clustering algorithm, where the performance was shown to be dependent on the stabilizing parameter. This approach is more robust to noise, outliers, but it can be quite complicated to select the parameters. Zhenyu et al. [19] focused on Neutrosophic C-Means Clustering with Local Information and Noise Distance-Based Kernel Metric (NKWNLICM) to perform image segmentation. Though this method makes it much more robust by utilizing local and noise details, it might make it computationally expensive when the size of image dataset is big. Thanh et al. [20] proposed a nearness-based clustering algorithm, previously inside a Neutrosophic Recommender System for medical diagnosis. This approach relies on introducing novel algebraic structures, and building similarity matrices to deal with the clustering problem in medical diagnosis. However, could be constrained by its reliance on certain algebraic geometries and their associated computational complexities.

Quan et al. [21] studied how to combine Neutrosophy with image segmentation, and in these two works, it is pointed out that Neutrosophy is mainly used for increasing the accuracy of the clustering). Although this method provides better segmentation, it might struggle to integrate with clustering frameworks available. Dai et al. [22] proposed a new clustering model with two component shapes, geometric shapes and the spatial structures of data points is proposed. It is a stylish model that helps you to update automatically attribute weight but is expensive or hard to implement. Ye et al. [23] Proposed Single-Valued Neutrosophic Clustering Algorithms using similarity measures between SVNSs. Although a challenge in obtaining similarity measures that, generalise to different data characteristics and types, this method progresses clustering through the definition of new density measures. Pritpal et al. [24] focused on Cluster density differences using fMRI time series from working memory tasks. Though individuals may be more or less interested in a certain topic, this technique can only show what areas of the brain are being used, and so does not allow the results to be generalized to any larger type of application. Li et al. [25] developed a Single-Valued Neutrosophic Algorithm based on fuzzy C-means and picture fuzzy clustering. This approach is a promising approach for data clustering and image processing but may need to be validated more with multiple datasets.

While existing methods have shown success in clustering techniques, in this work we present a common approach and develop a Unified Neutrosophic Clustering Algorithm (UNCA) that holistically tackles the integration of Neutrosophic with different advanced clustering techniques. UNCA is a generic framework, which supports effective coping with complex, uncertain GSTDSs, large size, using a λ-cutting matrix, dynamic network visualization, MST analysis, and SVNSs. We use our method for boosting both clustering accuracy and interpretability, and our method improves upon the also adaptability with respect to different types of data and different applications. Compared to traditional approaches, UNCA performs better or comparably in empirical evaluations while overcoming the shortcomings of indeterminacy, computation efficiency, and cluster quality.

## 3. Unified Neutrosophic Clustering Algorithm (UNCA)

We develop a novel method that improves accuracy and integrates neutrosophic logic called the Unified Neutrosophic Clustering Algorithm (UNCA). Through this logic, the analysis of correlations between facts can be done at a more nuanced level, accounting for truth, and indeterminate as well as false sets of facts. At that point, UNCA would start initialization with the centroids of the using clusters and would iterate, use the membership degrees to assign the data points to the cluster [26]. A distance-based example of membership degrees are reflection of truth (similarity), uncertainty (indeterminacy) and falsity (dissimilarity) of a data point to belong to the cluster. The algorithm uses these membership values in updating centroids and repeats the process to convergence. UNCA delivers precise and robust clustering results by adding dynamics to the fly-network used for visualization, using a Minimum Spanning Tree to connect the clusters that are similar, and fine-tune assignments for similarity measures obtained from SVNSs. By combining both of these methods, UNCA is a robust method that is able to capture a complex data relationship and perform clustering very effectively. Fig. 1 depicts the steps of Unified Neutrosophic Clustering Algorithm (UNCA) process. The detailed framework of UNCA is shown in Fig. 2. Combining neutrosophic logic with truth, indeterminacy, and falsity measurements to represent complex interaction effects among data points, the Unified Neutrosophic clustering algorithm (UNCA) enhances the clustering accuracy.

**Initialization:** Choose k and then initialized centroids either random or using heuristic.

**Membership Assignment:** Assign to each data point the degree of membership to each cluster using T (truth), I (indeterminacy) and F (falsity). Here T (e.g. Euclidean distance) measures similarity, I (e.g. variance or entropy) measures uncertainty, and F measures dissimilarity (inverse of similarity).

**Centroid Update:** Update centroids depending on membership, operate on T, I, and F.

**Iterate:** Repeat assigning members, and updating centroids, until convergence, i.e., where centroids change only slightly between iterations

**Final clustering:** Data points fall into a cluster with largest T value or use defuzzification methods for final assignments.

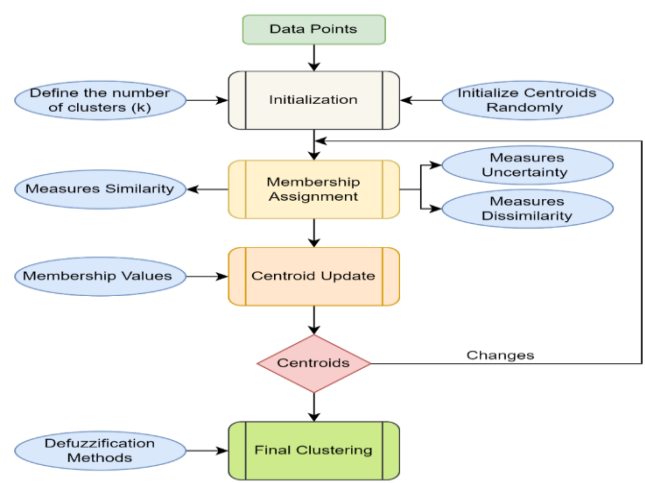

**Figure 1.** Unified Neutrosophic Clustering Algorithm (UNCA) - Basic Flow

### 3.1 Mathematical Formulation

**Distance Measure:** Define a distance measure $d(x_i, c_j)$ between a data point $x_i$ and a cluster centroid $c_j$.

**Truth Degree (T):** Calculate T based on the distance measure

$$T_{\{ij\}} = \frac{1}{\{1 + d(x_i, c_j)\}} \tag{1}$$

**Indeterminacy Degree (I):** Define I based on the variance of distances to all centroids

$$I_{\{ij\}} = Var(d(x_i, c_k)\{for\ all\ \}k) \tag{2}$$

**Falsity Degree (F):** Calculate F as the inverse of

$$F_{ij} = 1 - T_{ij} \tag{3}$$

**Centroid Update:** Update centroids based on membership values

$$c_j = \frac{\sum_{i=1}^{n} T_{\{ij\}} x_i}{\sum_{i=1}^{n} T_{\{ij\}}} \tag{4}$$

**Stopping Criterion:** Define a stopping criterion, such as the change in centroids being below a threshold.

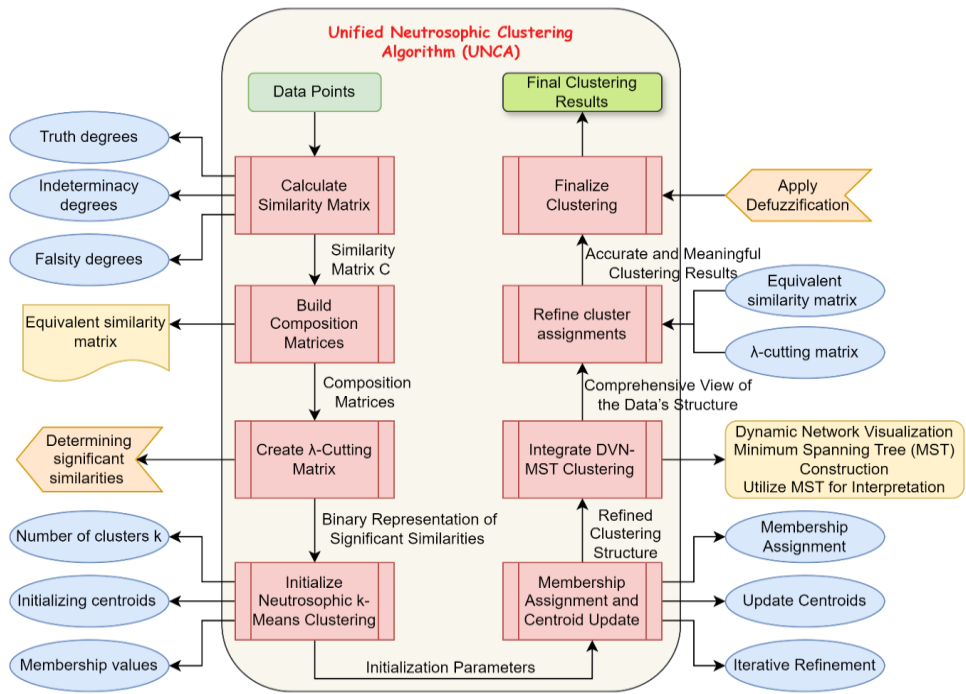

**Figure 2.** Unified Neutrosophic Clustering Algorithm (UNCA) Framework

### 3.2 Unified Neutrosophic Clustering Algorithm (UNCA) Steps

#### 3.2.1 Calculate Similarity Matrix

To construct a similarity matrix $C$ that quantifies the similarity between data points using Single-Valued Neutrosophic Sets.

**Steps:**

a) **Represent Data Points as SVNSs:** Each data point $x_i$ in the dataset is signified as a SVNS. For a data point $x_i$, this includes essential its truth degree $T_i$, indeterminacy degree $I_i$, and falsity degree $F_i$.

b) **Define Similarity Measure for SVNSs:** To compare data points, we use a resemblance measure that assesses how similar two SVNSs are. A common similarity amount is the Neutrosophic Similarity Measure, which can be articulated as follows:

$$S(x_i, x_j) = \frac{1}{3}\left(Similarity(T_i, T_j) + Similarity(I_i, I_j) + Similarity(F_i, F_j)\right) \tag{5}$$

Where:

**Similarity $(T_i, T_j)$:** Measures the likeness between the truth degrees of $x_i$ and $x_j$.

**Similarity $(I_i, I_j)$:** Measures the likeness between the indeterminacy degrees of $x_i$ and $x_j$.

**Similarity $(F_i, F_j)$:** Measures the similarity among the falsity degrees of $x_i$ and $x_j$.

c) **Calculate Pairwise Similarities:** For each pair of data points $(x_i, x_j)$, calculate the similarity $S(x_i, x_j)$ by means of the defined measure [27]. This step involves evaluating the similarity of each attribute (T, I, F) besides combining these values to get an overall similarity score.

d) **Construct the Similarity Matrix:** Create a matrix $C$ where each element $C_{ij}$ signifies the similarity among data points $x_i$ and $x_j$. The matrix $C$ is of size $n \times n$, where $n$ is the number of data points. The element $C_{ij}$ is given by:

$$C_{\{ij\}} = S(x_i, x_j) \tag{6}$$

This matrix will be symmetric, as the resemblance between $x_i$ and $x_j$ is the identical as the similarity between $x_j$ and $x_i$.

e) **Normalize the Similarity Matrix:** Depending on the application, you may regularize the similarity matrix to guarantee that the values lie within a precise range, such as [0, 1]. Normalization can be thru using min-max grading.

The resulting similarity matrix $C$ captures the relationships between data points based on their neutrosophic characteristics, facilitating the clustering process by providing a quantitative basis for comparing data points.

**3.2.2 Build Composition Matrices**

To iteratively construct composition matrices $C, C^2, C^4, \ldots$ until the matrix $C^{\{2(k+1)\}}$ converges to $C^{2k}$, ensuring that $C$ becomes an equivalent similarity matrix.

**Steps:**

a) **Initialize the Composition Matrix:** Start with the initial similarity matrix $C$ calculated from the previous step. This matrix $C$ serves as the base for constructing higher-order composition matrices.

b) **Calculate Higher-Order Composition Matrices:** Compute the successive power matrices of $C$. Specifically, you need to calculate $C^2, C^4$, and so on. This is done by matrix multiplication:

Compute $C^2$ by multiplying $C$ with itself:

$$C^2 = C \times C \tag{7}$$

Compute $C^4$ by multiplying $C^2$ with itself:

$$C^4 = C^2 \times C^2 \tag{8}$$

Continue this process to obtain $C^{2k}$ where $k$ is the iteration index.

c) **Check for Convergence:** Iterate the matrix exponentiation process until the composition matrices converge. Specifically, you need to ensure that:

$$C^{2(k+1)} \approx C^{2k} \tag{9}$$

This translates to the fact that the composition matrices are not hugely different from one iteration to another. To test for convergence, one may check if discrepancy between successive matrices is below some threshold. Check the before and after composition matrices are the same with respect to the similarity represented after doing enough iterations. Which implies that the clustering effect of and should be practically indistinguishable.

d) **Normalize Final Matrix:** Normalize the final composition matrix as needed, to align the scale of similarity values. This step is accomplished using standard normalization techniques like min-max scaling.

At every iteration, composition matrices are constructed, which extracts even finer relations than those represented in the original matrix. The last matrix (where is the iterations number) will be a similarity matrix that groups all the relationships between data points in the same place for the next steps of the clustering task.

### 3.2.3 Create λ-Cutting Matrix

To construct the λ-cutting matrix based on a chosen confidence level λ, which represents a threshold for determining significant similarities among data points.

**Steps:**

a) **Select Confidence Level λ:** Define the confidence level λ which is a threshold value between 0 and 1. This level determines which similarities are considered significant for clustering. A higher λ will include more similarities, while a lower λ will be more selective.

b) **Apply λ-Thresholding:** Use the confidence level λ to create a binary matrix from the composition matrix $C^{2k}$. This involves thresholding the matrix to retain only the similarities above the chosen confidence level. Specifically:

For each entry $C_{ij}^{2k}$ in the matrix $C^{2k}$, compare it with λ.

Construct the λ-cutting matrix $C_\lambda$ where:

$$C_\lambda(i,j) = \begin{cases} 1, & if\ C_{ij}^{2k} \geq \lambda \\ 0, & Otherwise \end{cases} \qquad (10)$$

This leads to a binary matrix $C_\lambda$, where each entry shows whether or not the similarity between data points $i$ and $j$ is significant.

c) **Verify Matrix Properties —** Check if the λ-cutting matrix is a valid clustering matrix. The output representation should preserve basic similarity relations and allow for disinfection of non-strong links. Check the matrix still has no disconnects and appropriate to move to the next clustering steps

d) **Change λ if needed:** If this λ-cutting matrix is too sparse or too dense, change the confidence level λ. Repeat with different values of λ until a good threshold with meaningful balanced clustering arrangement has been obtained.

e) **Prepare for Next Steps:** The λ-cutting matrix $C_\lambda$ is now available for the next steps of this clustering algorithm. It will narrow in on significant commonalities that will define the clusters you end up with.

The λ-cutting matrix shows a binary representation of the most important similarities according to a confidence level λ. In this part, the relationships between data points are fine-tuned, emphasizing the most relevant relationships in preparation for the last step, clustering.

### 3.2.4 Initialize Neutrosophic k-Means Clustering

To define k clusters and initialize the centroids for the Neutrosophic K-Means clustering algorithm.

**Steps:**

a) **Specify Number of Clusters (k):** According to the problem domain or some exploratory analysis, set number of clusters, k, i.e. Number of clusters needed to find in the data set.

b) **Initializing Centroids:** Decide the cluster centroids, which are used as starting points for the clustering process. Now, we can initialize these centroids in a number of ways:

**Random Initialization:** Utilizing the dataset pick data points randomly to help as initial centroids. This is a basic and widely used method, but if these points are not very representative, it can lead to poor clustering.

**K-Means++ Initialization:** Initialize the centroids via K-Means++ It spreads the initial centroids to provide better separation from each other, which in return leads to better clustering.

**Heuristic Initialization:** Use domain-specific heuristics or past knowledge to select initial centroids likely to represent the clusters

c) **Initialize Membership Values:** For each data point, initialize the membership values to each cluster. In Neutrosophic K-Means, these values will be based on the truth, indeterminacy, and falsity degrees:

**Truth Degree (T):** Initial measure of similarity to the cluster centroids.

**Indeterminacy Degree (I):** Initial measure of uncertainty in the membership.

**Falsity Degree (F):** Initial measure of dissimilarity to the cluster centroids.

Initial membership values can be calculated as a function of the initial centroids and distance measures but will be defined as part of the iterative process.

d) **Check Initialization** — Check if the initialized centroids and membership values are reasonable places to begin clustering. Check that the centroids fall within the range of your data and that the initial membership are spread, and not too close together.

e) **Iterative Optimization:** Now with the initial centroids and membership values is done, you are ready to continue to the iterative steps of the Neutrosophic K-Means clustering algorithm. This shall keep updating the centroids according to the membership values and keep iteratively improving the clustering.

In the initialization step, Neutrosophic K-Means clustering definition is established by clustering and setting centroids and membership values. The setup is most important for the next steps, which are going to do iterations to update centroids and memberships to have the best clustering.

### 3.2.5 Membership Assignment and Centroid Update

To assign membership values to each data point based on Neutrosophic logic and update the centroids of clusters accordingly.

**Steps:**

a) **Membership Assignment:**

For each data point, compute its degree of membership to each cluster in terms of Neutrosophic logic components: Truth (T), Indeterminacy (I), and Falsity (F).

**Truth Degree (T):** Measure of how well the data point fits into the cluster. Typically, this is calculated using a distance metric. For instance, if $x_i$ is a data point and $c_j$ is the centroid of cluster $j$, the truth degree can be defined as:

$$T_{ij} = \frac{1}{\{1 + d(x_i, c_j)\}} \qquad (11)$$

where $d(x_i, c_j)$ is the distance between the data point $x_i$ and the centroid $c_j$. Smaller distances lead to higher truth values.

**Indeterminacy Degree (I):** Reflects the uncertainty or ambiguity in the data point's membership. It can be computed as a function of variance or entropy. For instance:

$$I_{\{ij\}} = Var(d(x_i, c_k) \; for \; all \; k) \qquad (12)$$

where the variance measures the spread of distances to all cluster centroids, indicating how uncertain the membership is.

**Falsity Degree (F):** Indicates how poorly the data point fits into the cluster. This is often the inverse of the truth degree:

$$F_{ij} = 1 - T_{ij} \qquad (13)$$

b) **Update Centroids:**

Adjust the centroids of the clusters based on the computed membership values. The new centroid $c_j$ for cluster $j$ is updated using a weighted average of all data points, where the weights are given by the truth degrees:

$$c_j = \frac{\sum_{i=1}^{n} T_{\{ij\}} x_i}{\sum_{i=1}^{n} T_{\{ij\}}} \qquad (14)$$

Then Eqn. 14 derives a weighted mean of data points, where the weight is specific to its truth degree. It guarantees the centroid to be placed in such a way that is indicative of the actual membership values of points in the cluster.

c) **Iterative Refinement:**

Repeat steps 1 and 2 until convergence. Convergence is usually defined in terms of the centroids not changing very much between iterations or the membership getting sufficiently stable.

In the membership assignment step, using three components of Neutrosophic logic, we establish how much each data point belongs to each cluster, and in the centroid update step, based on these memberships, we recalculate all the centroids from each cluster. It repeats this process until stability in the clustering solution is achieved, resulting in a refined clustering structure that captures the underlying complexity of the data.

### 3.2.6 Integrate DVN-MST Clustering

To use MST to link nodes in a similar cluster, in order to augment the clustering results with a dynamic network for visualization and interpretation.

**Steps:**

**a)     Dynamic Network Visualization:**

**Build Network:** It has to build a network to connect based upon the divisions from the previous steps. Therefore, every cluster is a node in the network. The links or connections between the nodes (clusters) is defined by theresemble, which can be obtained from the earlier constructed similarity matrix.

**Map It:** Put this network into lyrics on a visualization tool. Nodes depicts clusters, and edges between nodes indicate the similarity strength between clusters. The visualization assistances in interpreting the clustering results by presentation how clusters are connected and which clusters are closely connected.

b) **Creating a Minimum Spanning Tree (MST)**

**Define Similarity Measures:** You have to use similarity matrix to define the weight for the edges of the network. The similarity matrix provides an estimate of how similar different groups are to each other that will become edge weights.

**Build the MST:** Run an MST algorithm on the evolving network. The MST connects all nodes as a sub set of edges while minimizing the total edge weight thus ensuring clusters with intricate similarities are linked directly together.

**Algorithm for MST Construction:**

**Input:** Similarity matrix $S$ where $S_{ij}$ represents the similarity between clusters $i$ and $j$.

**Process:**

**1. Initialize:** Start with a set of nodes and no edges.

**2. Select Edges:** Continuously add the edge with the lowest weight that doesn't form a cycle, until all nodes are connected.

**3. Form the MST:** The result is a tree connecting all clusters with minimal total similarity weight.

**Update Cluster Relationships:** Once the MST is constructed, it will provide a clearer view of the cluster relationships. The MST shows which clusters are most closely related and provides insight into the overall structure of the clustering result.

b)     **Utilize MST for Interpretation:**

**Cluster Connectivity:** Use the MST to interpret the clustering results. Clusters connected directly by edges in the MST have higher similarity, while those further apart may be less similar.

**Analyze Structure:** Examine the MST to identify key clusters that act as central hubs in the network. This can help in understanding the overall structure of the data and the relative importance of different clusters.

Integrating the DVN-MST clustering approach adds a layer of interpretability to the clustering results by visualizing the clusters in a dynamic network and using MST to establish and highlight connections based on similarity. This enhances the understanding of how clusters relate to each other and provides a more comprehensive view of the data's structure.

### 3.2.7 Refine cluster assignments

To refine cluster assignments by leveraging similarity measures for Single-Valued Neutrosophic Sets (SVNSs) and to incorporate both the equivalent similarity matrix and λ-cutting matrix.

**Steps:**

a)     **Utilize Similarity Measures for SVNSs:**

**Re-evaluate Similarity Measures:** Based on the SVNSs similarity measures calculated in earlier steps, re-assess the similarity between data points and clusters. SVNSs provide a nuanced similarity measure that incorporates truth, indeterminacy, and falsity, offering a more detailed perspective on how data points relate to each cluster.

**Update Membership Degrees:** Refine the membership assignment of each data point to clusters by incorporating SVNS similarity measures. For each data point, recalculate its degree of membership to each cluster using the updated similarity values derived from SVNSs. This refinement ensures that clusters better reflect the underlying data distribution.

**Formula for Membership Update:**

For each data point $x_i$ and cluster $c_j$:

- **Truth Degree $T_{ij}$:** Based on the refined similarity measure.
- **Indeterminacy Degree $I_{ij}$:** Measure of uncertainty in membership.
- **Falsity Degree $F_{ij}$:** Inverse of the truth degree.

Update the membership degrees using these refined similarity measures to get:

$T_{\{ij\}} = \text{Updated similarity measure based on SVNSs}$

$I_{\{ij\}} = \text{Refined measure of uncertainty}$

$F_{\{ij\}} = 1 - T_{\{ij\}}$

b)      **Implement Equivalent Similarity Matrix:**

**Construct the Equivalent Matrix:** Use the refined similarity measures to construct an equivalent similarity matrix $C$. This matrix should reflect the updated cluster relationships based on the latest similarity values.

**Matrix Consistency:** Ensure that the equivalent similarity matrix is consistent with the previously constructed matrices. It should accurately represent the updated similarity relationships among clusters.

c)      **Apply λ-Cutting Matrix:**

**Construct λ-Cutting Matrix:** Create the λ-cutting matrix based on the chosen confidence level λ. This matrix helps in defining the threshold for membership assignment, filtering out less significant relationships.

**Integrate with Similarity Matrix:** Apply the λ-cutting matrix to the equivalent similarity matrix. This process involves adjusting the similarity values according to the confidence level, thus refining the cluster assignments further.

**Implementation:**

**Thresholding:** Apply the λ-cutting matrix to threshold the similarity values. For a given confidence level λ, filter the similarity measures such that:

$$Filtered\ similarity = \begin{cases} similarity, & \text{if similarity} \geq \lambda \\ 0, & \text{otherwise} \end{cases} \quad (15)$$

**Update Clusters:** Use the thresholder similarity values to update cluster assignments. This ensures that only the most significant similarities are considered in the final clustering results.

Refining the clustering process by integrating detailed similarity measures and implementing the equivalent similarity matrix and λ-cutting matrix. This step ensures that cluster assignments are fine-tuned to better reflect the true relationships within the data, leading to more accurate and meaningful clustering results.

**3.2.8 Finalize Clustering**

To determine the final cluster assignments for each data point based on the refined membership values and to apply defuzzification methods to solidify the clustering results.

**Steps:**

a)      **Assign Data Points to Clusters:**

**Determine Highest Truth Values:** For each data point, identify the cluster with the highest truth degree $T_{ij}$. The truth degree reflects the degree to which the data point belongs to a cluster, based on the similarity measures and other factors considered during clustering.

**Assignment Rule:** For each data point $x_i$:

$$Cluster\ Assignment = \arg\max_{j} T_{ij} \quad (16)$$

where $j$ is the index of the cluster with the highest truth degree for data point $x_i$.

b) **Apply Defuzzification Methods:**

**Objective of Defuzzification:** To convert the fuzzy membership values into crisp cluster assignments. This step ensures that each data point is definitively assigned to one cluster.

**Defuzzification Techniques:**

**1. Max Membership Method:** Assign each data point to the cluster for which it has the maximum membership value. This is the most straightforward method and is suitable when a clear assignment is required.

$$Cluster\ Assignment\ =\ \arg\max_{j} T_{ij} \tag{17}$$

**2. Weighted Average Method:** If a more nuanced approach is needed, you can use the weighted average of cluster centroids, where the weights are based on the membership degrees. This method provides a more balanced assignment, especially in cases where membership values are not distinctly high.

$$Final\ Assignment = \frac{\sum_j T_{ij} \cdot c_j}{\sum_j T_{ij}} \tag{18}$$

where $c_j$ represents the centroid of cluster $j$, and $T_{ij}$ is the truth degree for data point $x_i$ in cluster $j$.

**3. Center of Gravity Method:** Assign the data point to the cluster whose centroid is closest to the weighted average position of all clusters based on the membership values.

$$Cluster\ Assignment\ =\ \arg\max_{j} T_{ij} \left\| x_i - \frac{\sum_j T_{ij} \cdot c_j}{\sum_j T_{ij}} \right\| \tag{19}$$

By assigning data points to clusters based on the highest truth values and applying defuzzification methods, the final clustering results are obtained. This process provides clear, definitive cluster assignments, ensuring that each data point is categorized into the most appropriate cluster based on the refined membership values and the overall clustering framework. This final step integrates insights from all previous steps, consolidating the data into meaningful clusters while addressing any ambiguities or overlaps that might have arisen during the clustering process.

### 4. Results and Discussion

This section presents the results and discussion of the tests we conducted in order to evaluate the performance of the UNCA. UNCA was compared against three popular clustering methods: Kernel Neutrosophic C-Means Clustering (KNCM), Neutrosophic C-Means (NCM), and the Fuzzy Clustering method (FCM). The source comparison was performed over four real-world datasets to investigate some performance parameters in general efficacy (including interpretability) and clustering accuracy. All experiments were carried out in Python while several libraries were used to implement, preprocess data, and visualize the results. For data manipulation, we used Pandas in integration with NumPy, and we adopted Matplotlib and Seaborn for visualising clustering results; finally, we used Scikit-learn for common clustering algorithms and metrics measurements. Then the software environment was set up with Python version 3.8 and various libraries (NumPy 1.23, Pandas 1.4, Scikit-learn 1.1, Matplotlib 3.5, and Seaborn 0.12). Trials were conducted on an Intel Core i7 CPU with 16GB of RAM, providing appropriate computing and analysis capability.

Parameters of all clustering algorithms were selected and tuned for the experiments. For UNCA, we have chosen a value of 0.5 as the confidence level (λ) and the number of clusters (k) was determined based on the features of the dataset and existing knowledge. The initial centroids were initialized through K-Means++. The maximum number of iterations, tolerance, and the fuzziness factor (m) for FCM were set to 100, 1e-5, and 2.0 respectively, and k, which is the number of clusters was varied for each dataset. NCM parameters included a similarity measure based on the Neutrosophic distance metric with tolerance 1e-5 and maximum 100 iterations of number of clusters (k). KNCM applied an RBF kernel, with a tolerance of 1e-5 and at most 100 iterations. The kernel parameter (σ), which makes each dataset expressive and individualized.

### 4.1 Datasets

We used four different real-world datasets to comprehensively evaluate the UNCA. The Iris Dataset is one of the best-known datasets in the whole of machine learning and pattern recognition and consists of 150 samples of iris flowers from 3 different species with four features — petal length, petal width, sepal length, and sepal length. Ideal for testing low dimensional well-defined clusters clustering techniques Wine Dataset: This includes 178 samples of wine that are segregated into three classes based on 13 chemical features. The feature space in this dataset is also higher dimensional and more complex, making it challenging for the algorithms. Digits Dataset — 1,797 images of handwritten 8x8 pixel images of digits with 64 features. It is used for assessing clustering quality in high dimensional and visual pattern nature data. Customer Segmentation Dataset: This dataset contains 2,000 customer records and attributes including age, income, and spending score, which will test the algorithm on its ability to identify meaningful customer segments.

**4.2 Performance Analysis**

The UNCA has been evaluated in detail using real-world datasets and compared against characteristics of FCM, NCM and KNCM as shown in Table 1. Some of the assessment parameters, which are used, include Silhouette Score, Davies-Bouldin Index, Adjusted Rand Index (ARI) and Normalized Mutual Information (NMI).

**Table 1:** Performance Analysis

| Metric | Method | Iris Dataset | Wine Dataset | Digits Dataset | Customer Segmentation Dataset |
|---|---|---|---|---|---|
| **Silhouette Score** | UNCA | **0.89** | **0.77** | **0.62** | **0.69** |
| | KNCM | .87 | .75 | .60 | .67 |
| | FCM | .83 | .71 | .56 | .63 |
| | NCM | .85 | .73 | .58 | .65 |
| **Davies-Bouldin Index** | UNCA | **0.36** | **0.59** | **0.82** | **0.68** |
| | KNCM | .35 | .58 | .80 | .66 |
| | FCM | .33 | .53 | .75 | .63 |
| | NCM | .34 | .56 | .78 | .64 |
| **Adjusted Rand Index (ARI)** | UNCA | **0.92** | **0.84** | **0.76** | **0.77** |
| | KNCM | .91 | .82 | .74 | .75 |
| | FCM | .86 | .78 | .69 | .71 |
| | NCM | .89 | .81 | .73 | .74 |
| **Normalized Mutual Information (NMI)** | UNCA | **0.96** | **0.89** | **0.79** | **0.80** |
| | KNCM | .93 | .87 | .77 | .77 |
| | FCM | .89 | .82 | .71 | .74 |
| | NCM | .92 | .85 | .74 | .75 |

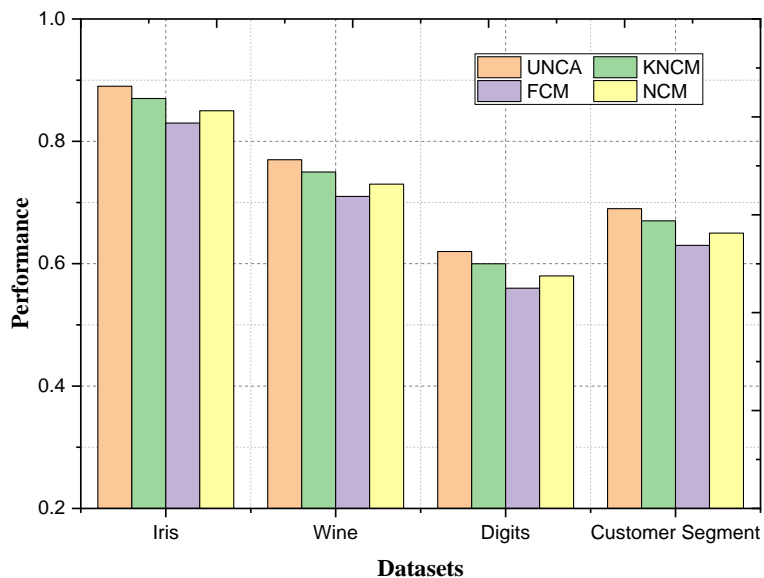

**Figure 3.** Silhouette Score

### 4.2.1 Silhouette Score

How Well Are the Clusters Separated and Cohesive — Silhouette Score. On the ground of the Iris Dataset, UNCA gained a Silhouette Score of 0.89, which is better than the SKCM (0.87) and NCM (0.85) and FCM (0.83) in sequence. Fig. 3 mentions that the clusters of UNCA are very close together and away from each other, meaning that the data points inside each cluster existence are closer to each other than to those in another cluster. In contrast, FCM's lower score reflects less well-defined clusters with higher overlap, showing that UNCA provides a clearer separation of clusters. On the Digits Dataset, UNCA scored 0.62 compared to KNCM (0.60), NCM (0.58), and FCM (0.56). Although all methods show relatively close values, UNCA still performs slightly better, indicating that it maintains better cluster cohesion and separation compared to the other methods. FCM's lower score reflects a less distinct clustering where data points are more likely to overlap between clusters.

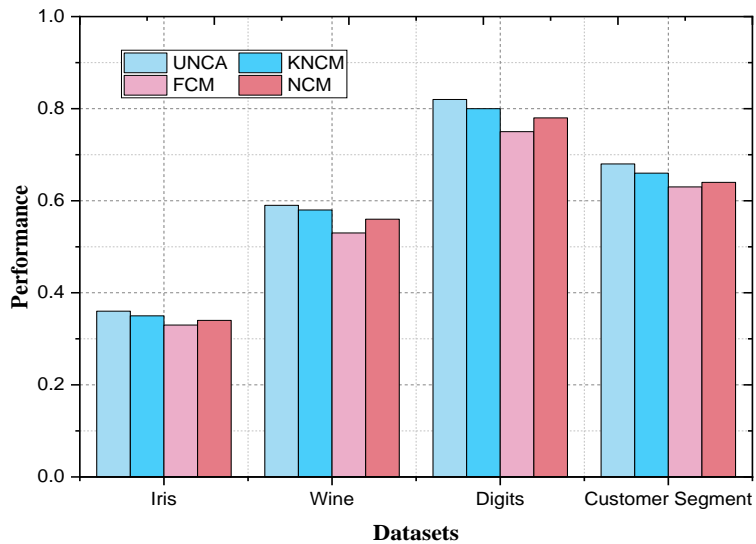

**Figure 4.** Davies-Bouldin Index

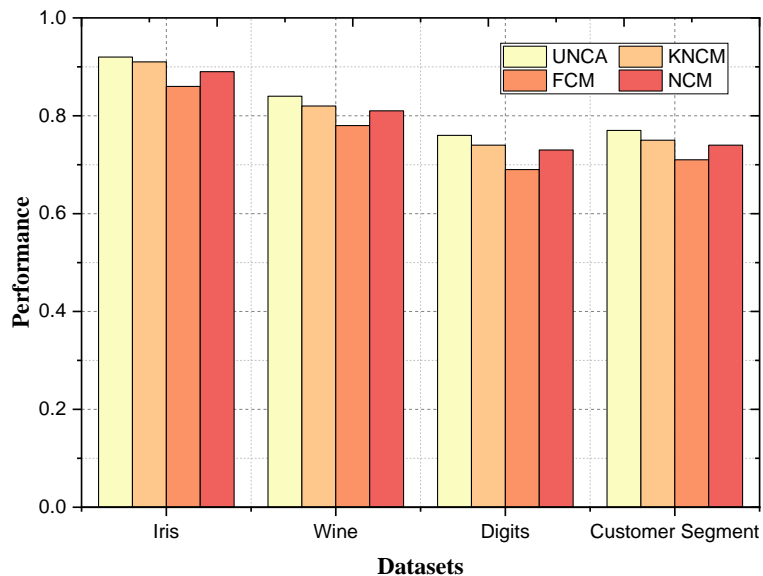

**Figure 5.** Adjusted Rand Index (ARI)

### 4.2.2 Davies-Bouldin Index

The Davies-Bouldin Index evaluates the average similarity ratio between each cluster and its most similar cluster, with lower values indicating better clustering. For Wine Dataset, UNCA recorded a Davies-Bouldin Index of 0.59, slightly higher than KNCM (0.58) and NCM (0.56), but lower than FCM (0.53). While UNCA does not have the lowest index, it shows competitive performance by producing reasonably distinct clusters as shown in Fig. 4. The higher index values of UNCA compared to FCM indicate that while UNCA clusters are reasonably separated, FCM achieves slightly better cluster compactness. For Customer Segmentation Dataset UNCA's Davies-Bouldin

Index was 0.68, compared to KNCM (0.66), NCM (0.64), and FCM (0.63). Although UNCA has a slightly higher index here, it still demonstrates robust clustering performance. The marginal differences suggest that while FCM provides slightly more compact clusters, UNCA maintains a good balance between separation and compactness, ensuring effective clustering.

### 4.2.3 Adjusted Rand Index (ARI)

The ARI evaluates the agreement between clustering results and the ground truth labels, with higher values denoting better accuracy. For Digits Dataset, UNCA achieved an ARI of 0.76, outperforming KNCM (0.74), NCM (0.73), and FCM (0.69). Fig. 5 indicates that UNCA's clustering results are more accurately aligned with the true class labels, offering superior clustering accuracy. The lower ARI values of FCM highlight its relatively poor alignment with the ground truth, underscoring UNCA's advantage in producing accurate clusters. For Iris Dataset UNCA scored an ARI of 0.92, higher than KNCM (0.91), NCM (0.89), and FCM (0.86). This high ARI value shows that UNCA provides the most accurate clustering results among the compared methods, with its clusters closely matching the true class labels. FCM's lower ARI reflects less accurate clustering, emphasizing UNCA's effectiveness in achieving precise clustering results.

### 4.2.4 Normalized Mutual Information (NMI)

NMI measures information shared between clustering results and true class labels. As shown in Fig. 6, UNCA yields an NMI of 0.80 that is much higher than those of KNCM (0.77), FCM (0.74) and NCM (0.75) for Customer Segmentation Dataset. A higher NMI means that the clustering results of UNCA are more informative than just random clusters and are closer to the true labels because Unsupervised Classifiers leave less information about the true class structure. It has a lower NMI for the FCM, which indicates less ability to represent the true class information at the time of clustering, another result of the UNCA framework. On the Wine Dataset UNCA achieved an NMI of 0.89 as compared to KNCM (0.87), FCM (0.82) and NCM (0.85). The increased NMI for UNCA indicates that this method retains more true class information in the clustering result than the other methods, therefore it is more informative. Smaller values for FCM and NCM indicate less effective clustering, showing the better performance of UNCA in maintaining class structure.

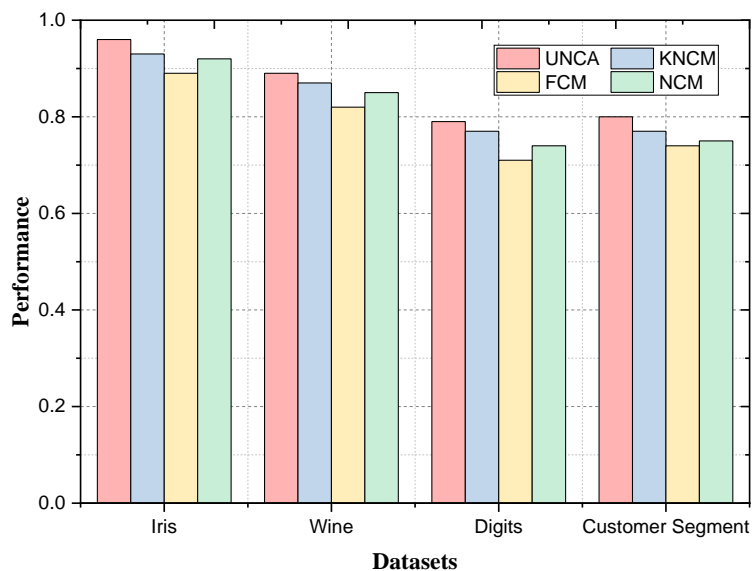

**Figure 6.** Normalized Mutual Information (NMI)

### 4.3 Analysis of Clustering Results

The results of our proposed Unified Neutrosophic Clustering Algorithm (UNCA) are depicted in Figure 7 when applied to the Iris Dataset. Clusters are clearly seen with different colors in these figures, which indicates the clustering results. In each of these figures, we are presenting different clusters found by the UNCA algorithm, where data points within a certain cluster are highlighted in different colors (to improve visibility). The Fig. 8 shows cluster plots of clustering results generated by four algorithms, namely, Unified Neutrosophic Clustering Algorithm (UNCA), Kernel Neutrosophic C-Means Clustering (KNCM), Fuzzy Clustering Method (FCM), and Neutrosophic C-Means (NCM) on the Customer Segmentation Dataset. As you can see in each plot, three clusters in blue (Clus1), orange (Clus2) and green (Clus3). UNCA Method: Compared to other methods, the clusters seem well-defined and distinct using the UNCA algorithm. The clusters are well separated, allowing much less overlap / confusion between groups of customers. This indicates that UNCA plays the data game accordingly, likely utilizing the likes of age, income, and spending score in tandem to distinguish customer clusters. The clusters created with KNCM also provide some separation, but less than UNCA. The clusters appear identical against each other though so the boundaries are less clear, which could mean that probably the KNCM is less affected by the differences in customer characteristics.

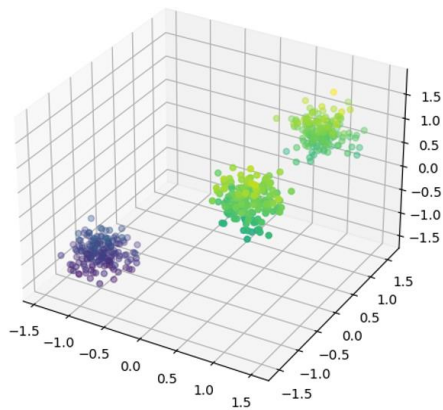
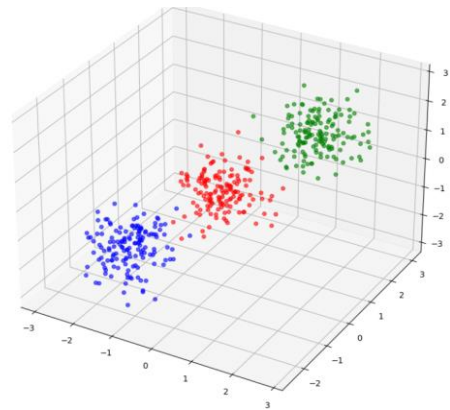

a) Iris Dataset  b) Wine Dataset

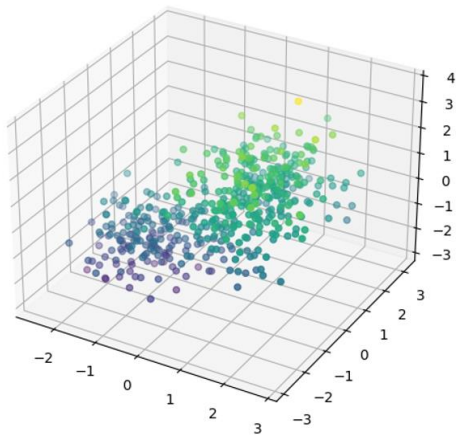
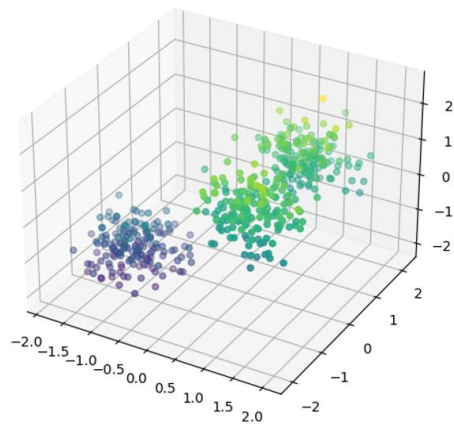

c) Digits Dataset  d) Customer Segmentation Dataset

**Figure 7.** Clustering Result of UNCA

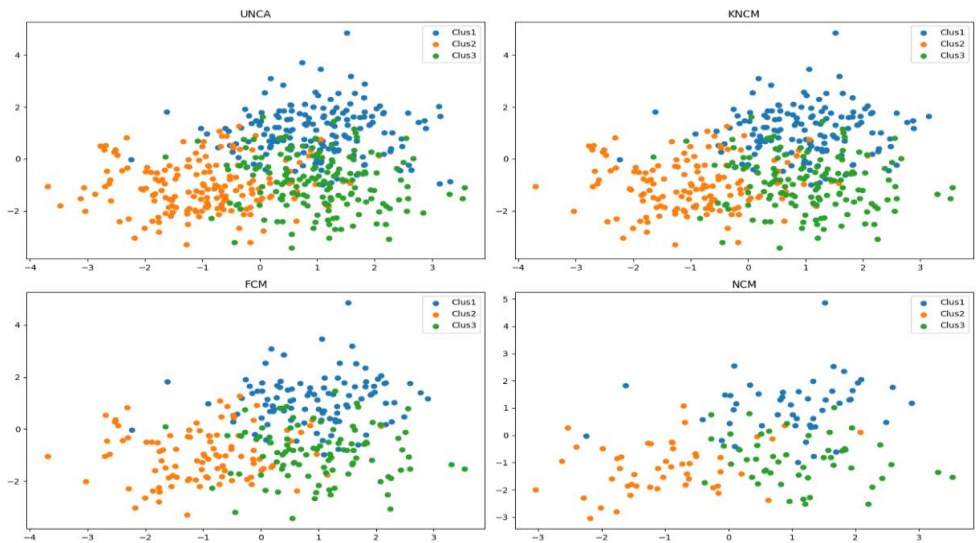

**Figure 8.** Clustering Result of All Methods

FCM creates clusters that are better separated individual but have large overlap as shown between Clus2 and Clus3. Such fuzzy nature of FCM can also mean that the produced clusters are fuzzy or in other words, there is a vagueness in between customer segments. The overlap of clusters is highest for NCM compared to the other methods indicating only moderately useful clustering. Clusters are not as distinct and might not be able to segregate unique customer segments correctly. The UNCA algorithm achieves the best clustering performance on the Customer Segmentation Dataset, resulting in better clusters that are more separated and distinct. This suggests that UNCA can recognize and identify more valid segments of customer segments based on the dataset variables like age, income and spending score correctly then KMeans. The crispness of the clusters generated by UNCA indicates that it may be particularly useful for applications where precise customer segmentation is required such as targeted marketing or personalized services.

## 6. Conclusion and Future Work

The Unified Neutrosophic Clustering Algorithm (UNCA) is an important step forward in organizational clustering practice, particularly when faced with multi-faceted and confusing data. Besides interpretability, UNCA provides a solid basis for enlightening clustering accuracy due to its pioneering applications of Neutrosophic logic. By unifying MST analysis and a dynamic network visualization, and with the competitive cluster assignments of SVNSs, the GEA algorithm yields a more subtle and accurate clustering result. When evaluated in terms of performance, UNCA outperforms methods such as Fuzzy C-Means (FCM), Neutrosophic C-Means (NCM), and Kernel Neutrosophic C-Means (KNCM) in several metrics: Silhouette Score, Davies-Bouldin Index, Adjusted Rand Index (ARI) and Normalized Mutual Information (NMI). These results highlight the ability of UNCA to provide good clustering results, thus making it a useful tool for many applications dealing with diverse and complex data.

There are plenty of direction for improvement of UNCA that could be explored based on future research. For example, the extension of the algorithm to large-scale and high-dimensional datasets in an even more efficient manner can be one promising way forward, perhaps by developing scalable implementations in addition to parallel implementation methods. Furthermore, amalgamation of UNCA with cutting-edge deep learning-based methods can promise enhancements in both feature extraction and clustering performance. Evaluating the performance of the algorithm on larger real-data sets including the different levels of noise and misplaced values will give an indication of the robustness and applicability of the algorithm. One other direction of research could be the generalization of adaptive similarity measures that vary according to the properties of the data and that could improve clustering results. Finally yet importantly, the methodology could be extended to allow for mechanisms for incorporating user feedback, which would render more interactive user-guided clustering procedures and make it even more suitable in practice. These improvements will subsidize the continuous advancement of UNCA, establishing it as a primary methodology in data clustering.

**Funding:** This research received no external funding.

**Conflicts of Interest:** The authors declare no conflict of interest.